\documentclass{article}
\usepackage{spconf,amsmath,epsfig}

\usepackage[]{graphicx}
\usepackage{caption}
\usepackage{amsmath}
\usepackage{amssymb}
\usepackage{epsfig}
\usepackage{cite}
\usepackage{color}
\usepackage{balance}
\usepackage{amsmath}
\usepackage{amsfonts} % for the \checkmark command 
\usepackage{graphicx}
\usepackage{pdfpages}
\usepackage[T1]{fontenc} % For less than and greater than sign
\usepackage{array}
\usepackage{url}

\usepackage{color}
\usepackage{wrapfig}
\usepackage[hidelinks,bookmarks=false]{hyperref}
\usepackage{multirow}
\usepackage{tabularx}
\usepackage{lipsum}
\usepackage{algpseudocode} % For algorithm 
\usepackage{algorithm,algpseudocode}
\usepackage{stackengine}
\usepackage{subcaption}

\begin{document}

\title{Short-term prediction of localized cloud motion\\using ground-based sky imagers}

\name{Soumyabrata Dev$^{1}$, Florian M. Savoy$^{2}$, Yee Hui Lee$^{1}$, Stefan Winkler$^{2}$\thanks{This research is funded by the Defence Science and Technology Agency (DSTA), Singapore.}\thanks{Send correspondence to \url{Stefan.Winkler@adsc.com.sg.}}}
\address{
	$^{1}$~School of Electrical and Electronic Engineering, Nanyang Technological University (NTU), Singapore \\  
	$^{2}\,$Advanced Digital Sciences Center (ADSC), University of Illinois at Urbana-Champaign, Singapore \\
}

\maketitle

\begin{abstract}
Fine-scale short-term cloud motion prediction is needed for several applications, including solar energy generation and satellite communications. In tropical regions such as Singapore, clouds are mostly formed by convection; they are very localized, and evolve quickly. 
We capture hemispherical images of the sky at regular intervals of time using ground-based cameras. They provide a high resolution and localized cloud images. We use two successive frames to compute optical flow and predict the future location of clouds. We achieve good prediction accuracy for a lead time of up to $5$ minutes.
\end{abstract}

\section{Introduction}
\label{sec:intro}

Cloud tracking aims at predicting the sky/cloud condition with a certain lead time. In solar energy generation, a forecast of the amount of sun light reaching a solar panel is needed to take preventive actions before a drop in energy output. Similarly, an accurate forecast of cloud movement helps in satellite communication systems to switch to a different ground station when the link is in danger of being affected by clouds \cite{yeo2011performance}.

Recently, ground-based sky cameras are increasingly used by remote sensing analysts to study the earth's atmosphere~\cite{GRSM2016}. These cameras are popularly known as Whole Sky Imagers (WSIs). In our research, we build our own WSIs, which we call WAHRSIS~\cite{WAHRSIS,IGARSS2015}. WAHRSIS stands for Wide-Angle High-Resolution Sky Imaging System. They consist of a DSLR camera with a fish-eye lens controlled by a single-board computer in a weather-proof box with a transparent dome, capturing hemispheric images of the sky. We currently have three imagers installed on rooftops of various buildings at the Nanyang Technological University campus. These imagers capture images at intervals of $2$ minutes and archive them in a server.

We use a sequence of images taken by those imagers to create a forecast with a lead time from $2$ up to $10$ minutes using optical flow. This is a challenging task, as clouds continuously change their shape and size. The achievable lead time is also restricted because of the limited field of view of the sky imager; depending on wind conditionds, coulds may move out of the field of view rather quickly.

The structure of this paper is as follows: we discuss the related work in Section~\ref{sec:rworks} and the method proposed in Section~\ref{sec:method}. Experimental results are presented in Section~\ref{sec:results}. Section~\ref{sec:concl} concludes the paper.

\section{Related Work}
\label{sec:rworks}
Traditionally, cloud tracking has been performed from satellite images for accurate weather prediction. Pioneer work of tracking involved detecting optical flow patterns in satellite images~\cite{trackingpioneer}. These flow patterns are useful to detect the evolution of several weather patterns. Sieglaff et al.\ \cite{Sieglaff13} fused data from satellites images, radar, and numerical models to understand the evolution of convective clouds. Recently, ground-based sky cameras are increasingly used for the purpose of tracking clouds in a localized manner. Porter and Cao \cite{Porter09} used stereo cameras to estimate the wind speed and its direction in the troposphere. Very recently, estimation of cloud motion from sky cameras has been used in solar irradiance forecasting~\cite{Chow15,Chauvin2016}. 

\section{Methodology}
\label{sec:method}
In this section, we first give a general formulation of the optical flow technique and then describe how it is applied to our problem.

\subsection{Optical Flow Formulation}

Optical flow is based on the brightness constancy constraint, which states that pixels of an image sequence do not change value, but only shift position over time. Let's define a pixel intensity at image coordinates $(x,y)$ and time $t$ by $I(x,y,t)$. Under this condition, this intensity would have moved by $(\Delta x, \Delta y)$ after a time $\Delta t$:
\[I(x,y,t) = I(x + \Delta x, y + \Delta y, t + \Delta t).\]

Using a first order Taylor series expansion, the optical flow equations can be derived \cite{fleet2006optical}:
\[\frac{\partial I}{\partial x}u_x + \frac{\partial I}{\partial y}u_y + \frac{\partial I}{\partial t} = 0,\]
where $(u_x, u_y)$ is the optical flow (or velocity), and $\frac{\partial I}{\partial x}$, $\frac{\partial I}{\partial y}$ and $\frac{\partial I}{\partial t}$ are the derivatives of the image in the $x$, $y$, and $t$ dimensions.

This equation cannot be solved analytically, as it has two unknowns. This is known as the \emph{aperture problem}. In order to solve it, two main approaches exists:
\begin{itemize}
\item Local methods state that the optical flow vectors are constant within some neighborhood and thus increase the number of equations to solve for the same optical flow vector. A typical example is the Lucas-Kanade method \cite{lucas1981iterative}.
\item Global methods assume that the optical flow vector distribution should be smooth across the spatial and temporal axes. They mimimize a global energy function in order to reduce large optical flow gradients. A famous algorithm is the Horn-Schunck method \cite{horn1981determining}.
\end{itemize}

We use the implementation from~\cite{liu2009beyond}\footnote{Available at \url{https://people.csail.mit.edu/celiu/OpticalFlow/}}, which is based on a combination of both local and global approaches, following the method proposed in~\cite{bruhn2005lucas}.

\subsection{Cloud Tracking}
We use two image frames taken at times $t-2$ minutes and $t$, and compute the translational vectors in both $x$ and $y$ direction of the images, using the method mentioned above. We rely on the assumption that clouds do not significantly change between image frames. The optical flow vectors estimate the direction and orientation of the moving clouds, assuming an affine transformation between two frames. %that a cloud captured at two different time frames only differ by a translation. 
The algorithm provides a dense result, i.e.\ a velocity vector is associated to every pixel coordinate in the input image.

Since we are interested in tracking the detailed cloud shape, it is important to use a color space which provide a good separation of clouds and sky. We use a variant of the ratio of red and blue color channels of the image. In an earlier work~\cite{ICIP1_2014}, we have analyzed various color channels and concluded that $(B-R)/(B+R)$ is the most discriminatory color channel, where $B$ and $R$ indicate the blue and red color channels respectively.

\begin{figure}[htb]
\centering
\begin{subfigure}[t]{0.22\textwidth}
\includegraphics[width=\textwidth]{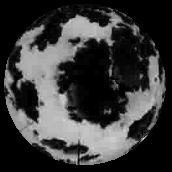}
\caption{Frame at time \emph{t-2} minutes}
\end{subfigure}\,
\begin{subfigure}[t]{0.22\textwidth}
\includegraphics[width=\textwidth]{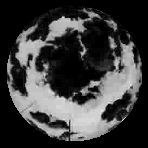}
\caption{Frame at time \emph{t}}
\end{subfigure}
\caption{Ratio channels of two successive image frames.
\label{fig:ratioimage}}
\end{figure}

As an illustration, we show the ratio channel of two successive image frames in Fig.~\ref{fig:ratioimage}. We observe that there is a clear contrast in this channel. These ratio channels are used to estimate the flow field of clouds.

\begin{figure}[htb]
\begin{center}
\stackunder[5pt]{\includegraphics[height=0.15\textwidth]{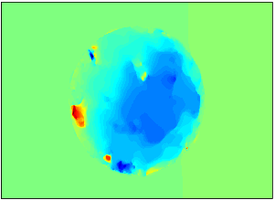}}{\small (a) Horizontal translation}
\includegraphics[width=0.042\textwidth]{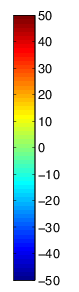}
\stackunder[5pt]{\includegraphics[height=0.15\textwidth]{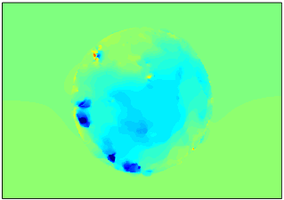}}{\small (b) Vertical translation}
\caption{Horizontal and vertical translation of pixels between Frame 1 and Frame 2. The color map represents the speed in pixels/minute units.
\label{fig:vectorflow}}
\end{center}
\end{figure}

Using the images in Fig.~\ref{fig:ratioimage}, we obtain the horizontal and vertical translations of each of the pixels, which can be combined into a vector field. We show them in Fig.~\ref{fig:vectorflow}, plotted in terms of pixels/minute. This provides an idea about the relative speed and direction of each of the pixels between the two frames.

\subsection{Cloud Motion Prediction}
We can now use the above information to predict future frames. The vector fields are applied individually to each of the red, green, and blue channels of the image.

Clouds have an ill-defined shape, and can change their shape and size very quickly. However, even though the underlying assumptions do not exactly match this reality, they work well to track clouds for short lead times, as is the case in many other optical flow problems.

For higher lead times, we use the actual frame at time $t$ and predicted frame at time $t+2$ minutes to predict the frame at time $t+4$ minutes. Similarly, we use the  predicted frame at time $t+2$ minutes and the predicted frame at $t+4$ minutes, to predict the frame at time $t+6$ minutes, and so on. At every stage, we use the previous two frames (actual or predicted), to compute the subsequent frame. This works well under the assumption that the clouds do not \emph{significantly} change their shape and location for the given lead time. 

\begin{figure*}[htb]
	\centering
    \begin{subfigure}[t]{0.13\linewidth}
        \centering
        \includegraphics[trim={6.9cm 2cm 7cm 1.8cm},clip,width=2.21cm]{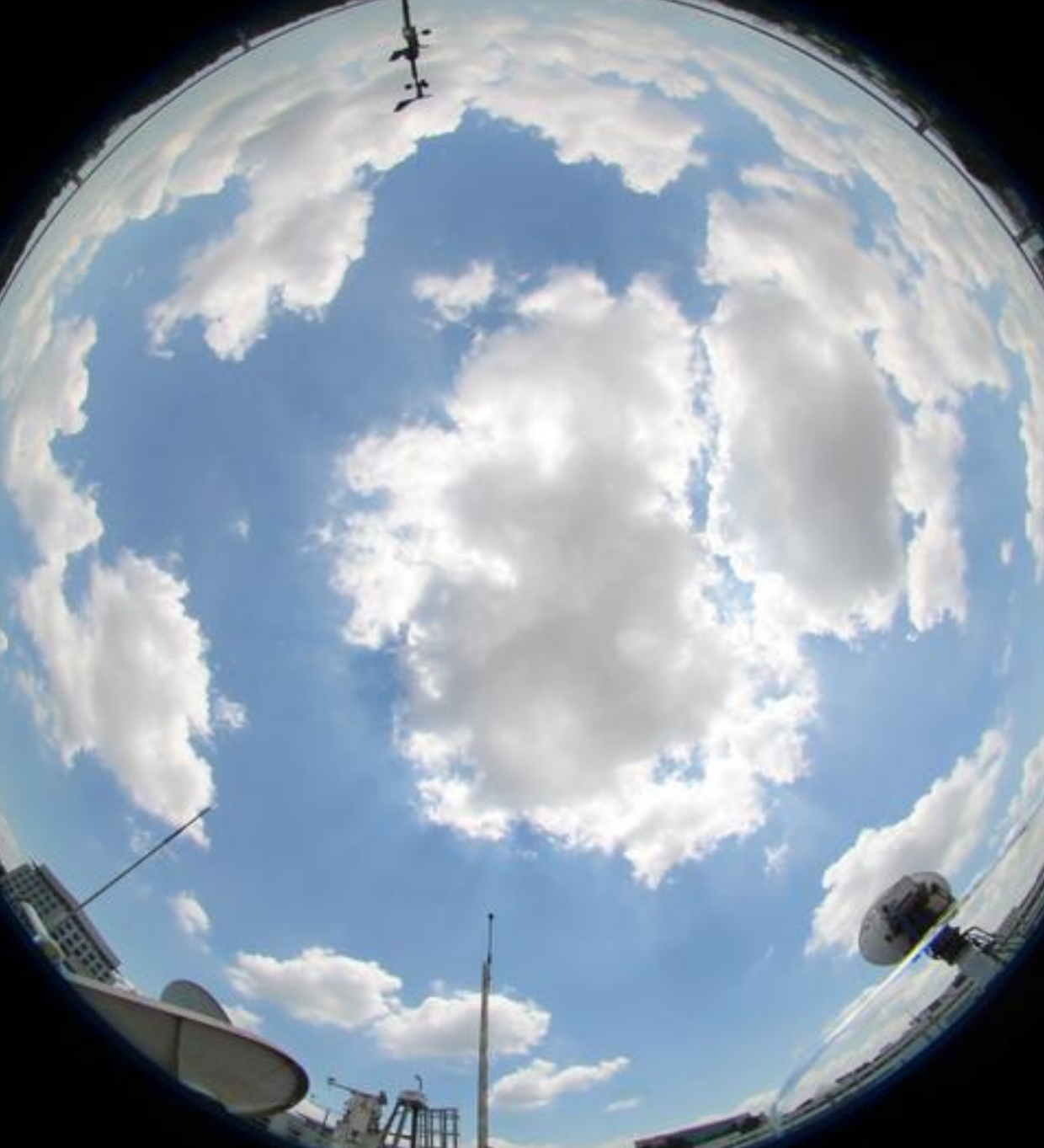}\\[2mm]
        \includegraphics[trim={6.9cm 2cm 7cm 1.8cm},clip,width=2.21cm]{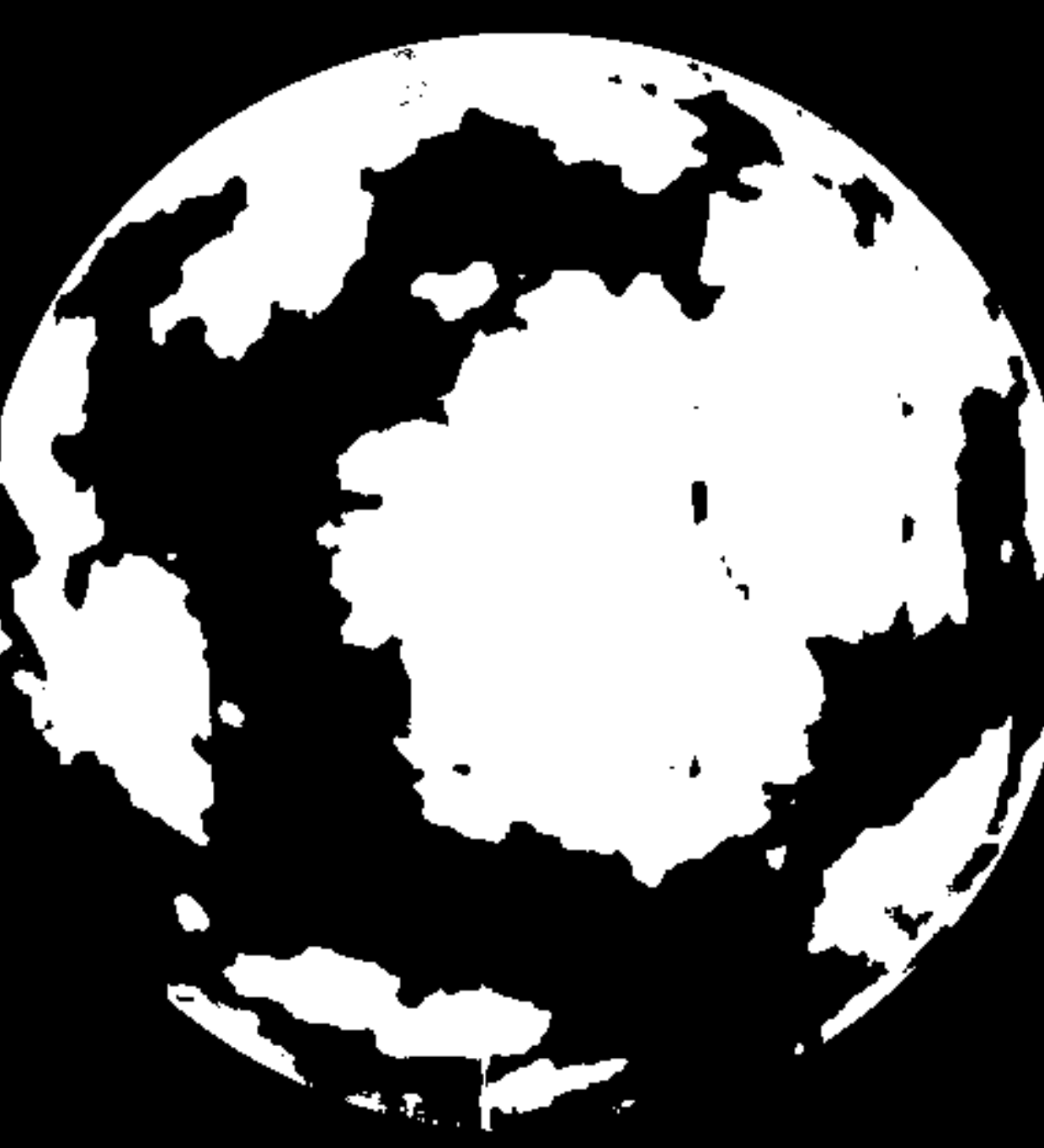}
        \caption{Input at $t-2'$}
    \end{subfigure}
    \begin{subfigure}[t]{0.13\linewidth}
        \centering
        \includegraphics[trim={6.9cm 2cm 7cm 1.8cm},clip,width=2.21cm]{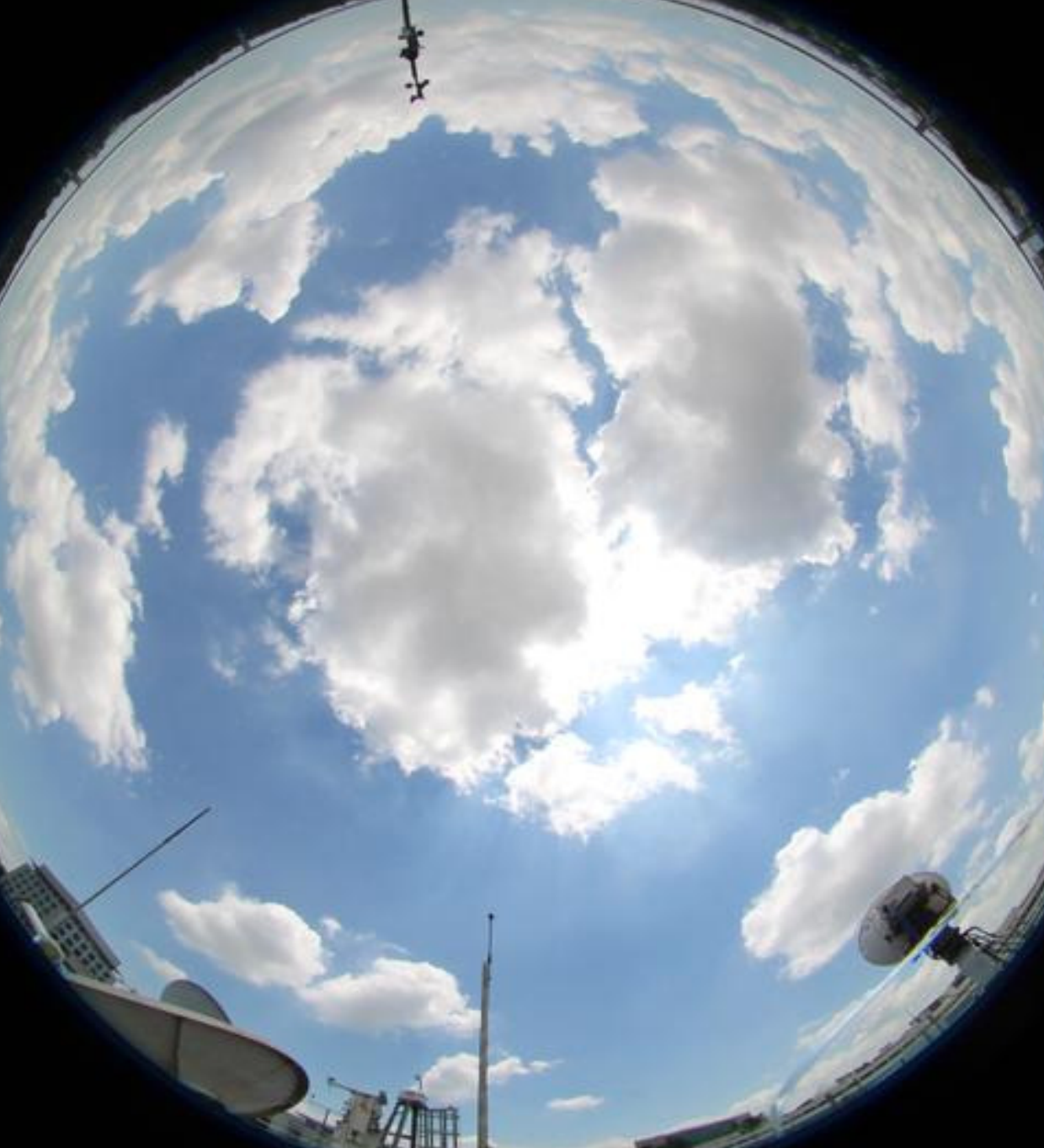}\\[2mm]
        \includegraphics[trim={6.9cm 2cm 7cm 1.8cm},clip,width=2.21cm]{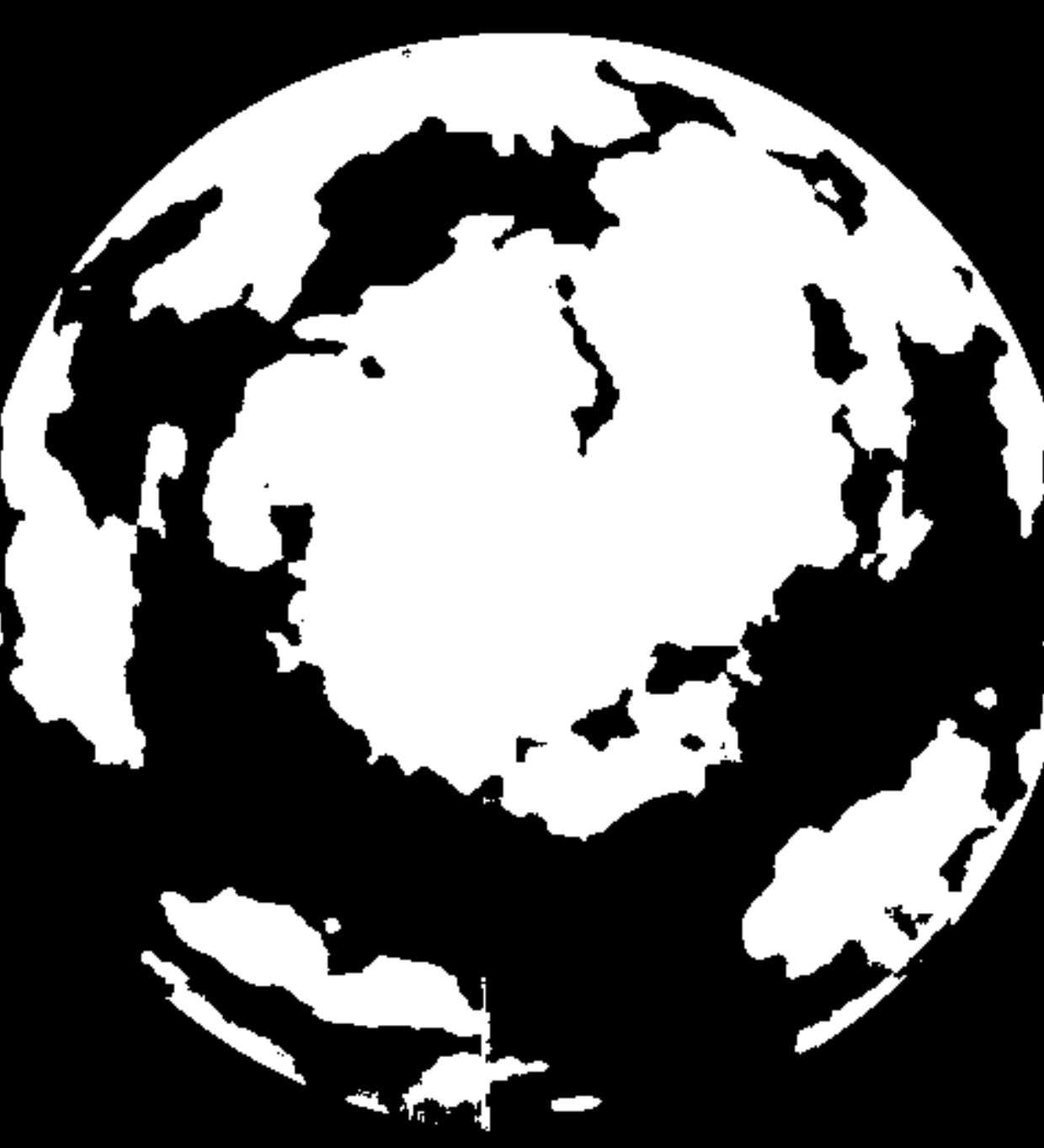}
        \caption{Input at $t$}
    \end{subfigure}
    \hspace{0.2cm}
    \begin{minipage}[t]{0.02\linewidth}
    	\vspace{-1.4cm}
    	\begin{minipage}[t][1cm]{\textwidth}
        	\rotatebox{90}{Actual}
        \end{minipage}\\[8mm]
        \begin{minipage}[t][1.2cm]{\textwidth}
            \rotatebox{90}{Actual binary}
        \end{minipage}\\[5mm]
        \begin{minipage}[t][1.2cm]{\textwidth}
            \rotatebox{90}{Predicted}
        \end{minipage}\\[6mm]
        \begin{minipage}[t][1.2cm]{\textwidth}
            \rotatebox{90}{Predicted binary}
        \end{minipage}
    \end{minipage}
    \begin{subfigure}[t]{0.13\linewidth}
        \centering
        \includegraphics[trim={6.9cm 2cm 7cm 1.8cm},clip,width=2.21cm]{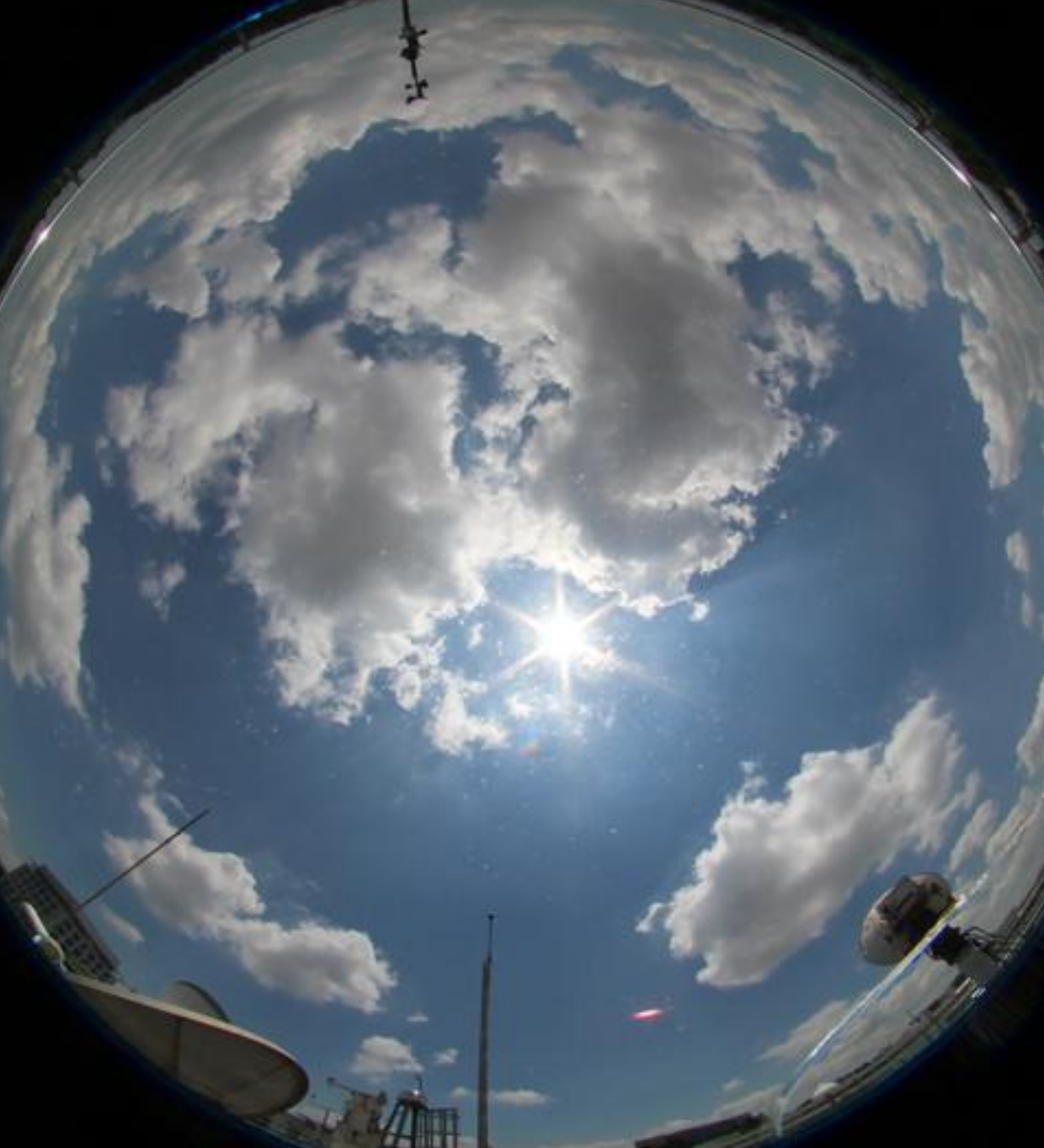}\\[2mm]
        \includegraphics[trim={6.9cm 2cm 7cm 1.8cm},clip,width=2.21cm]{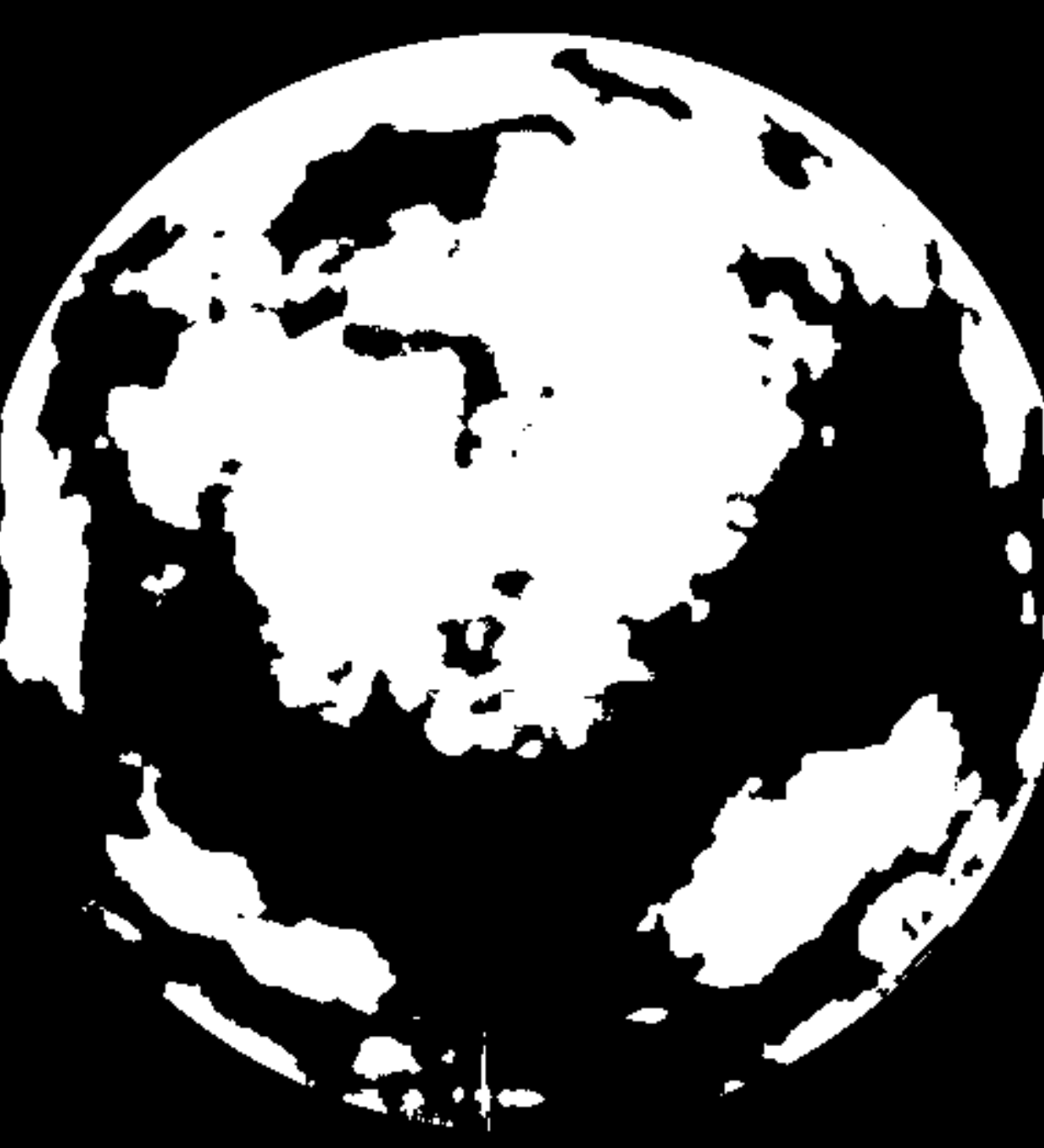}\\[2mm]
        \includegraphics[trim={6.9cm 2cm 7cm 1.8cm},clip,width=2.21cm]{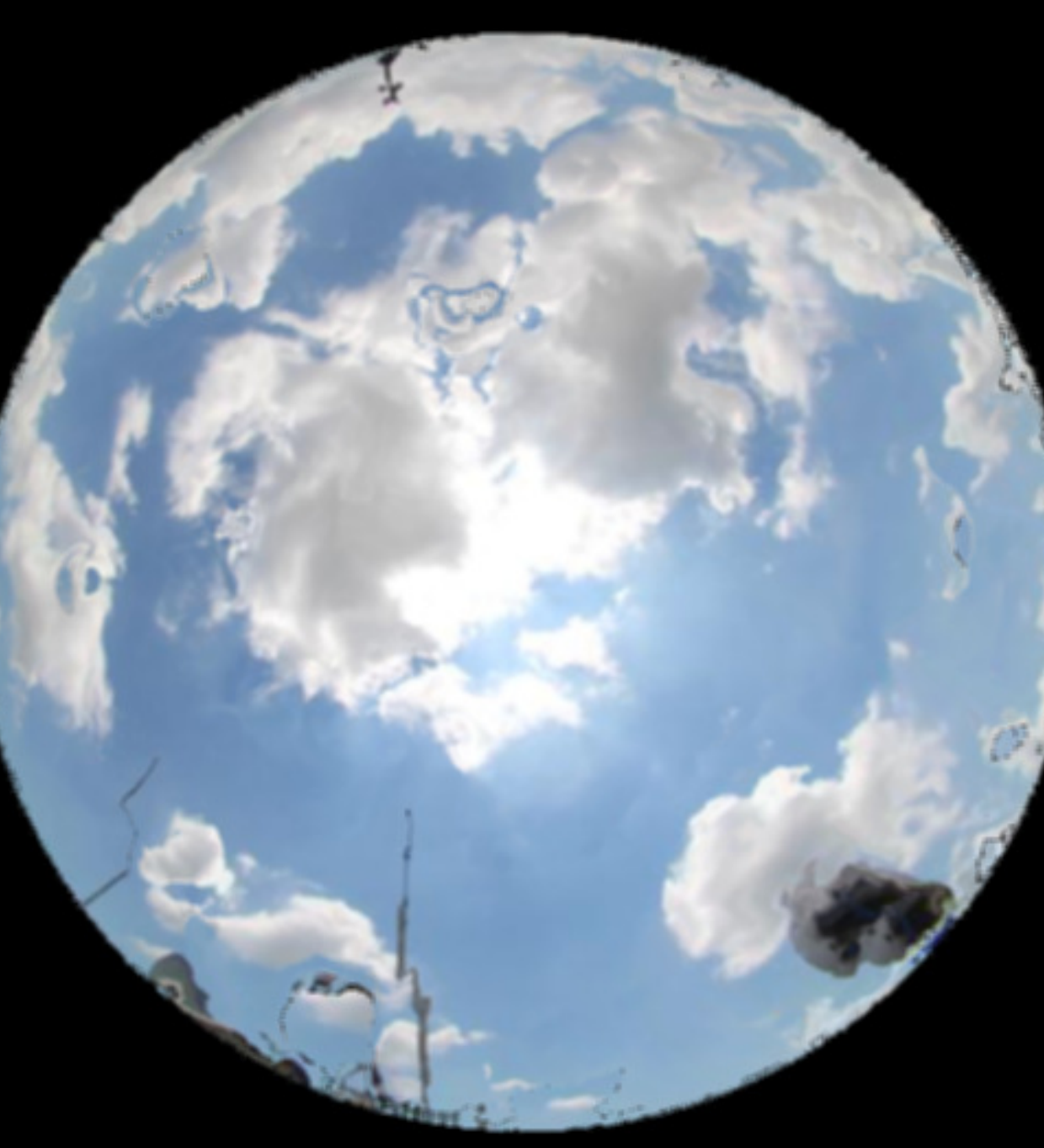}\\[2mm]
        \includegraphics[trim={6.9cm 2cm 7cm 1.8cm},clip,width=2.21cm]{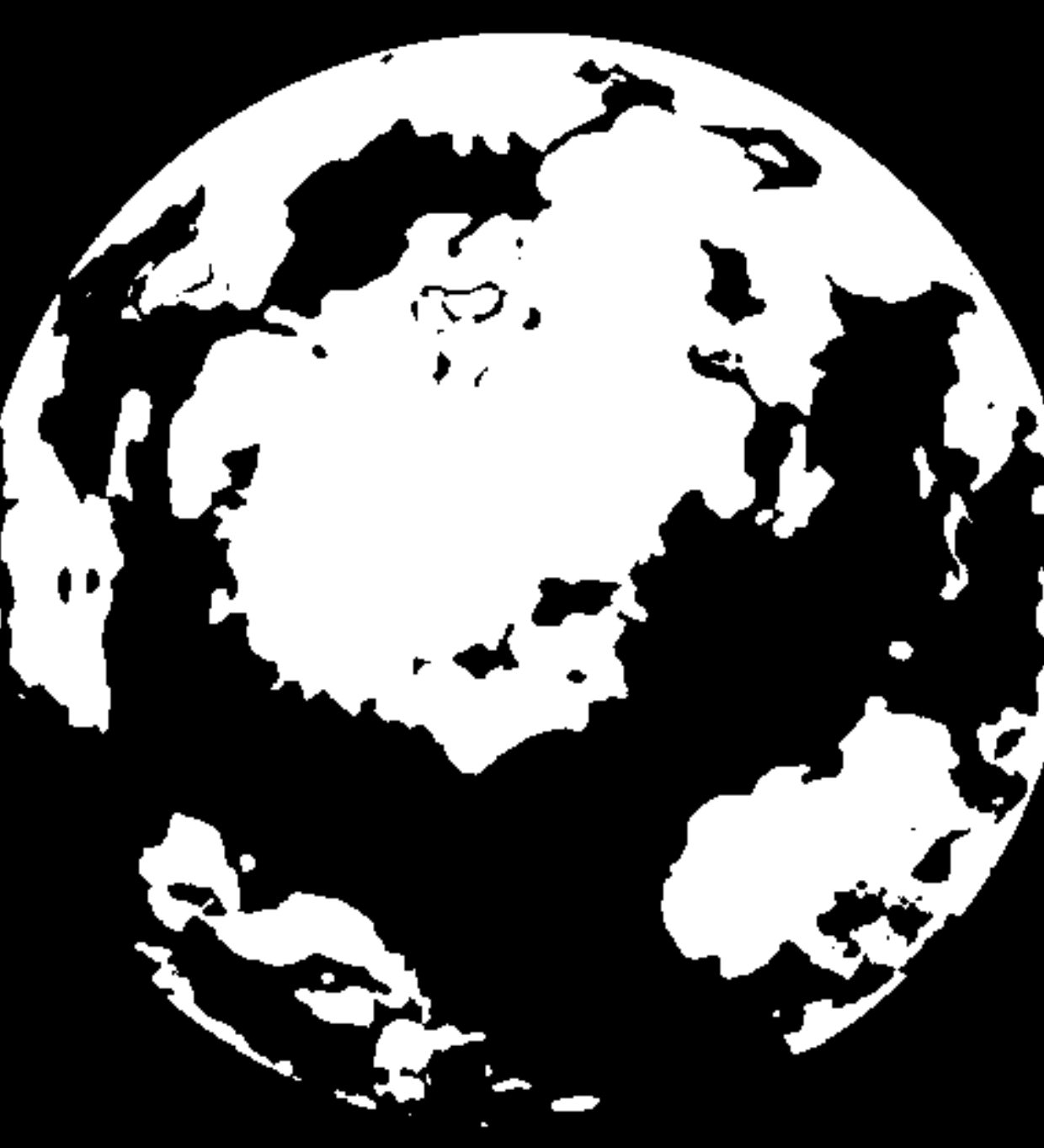}
        \caption{$t+2'$}
    \end{subfigure}
    \begin{subfigure}[t]{0.13\linewidth}
        \centering
        \includegraphics[trim={6.9cm 2cm 7cm 1.8cm},clip,width=2.21cm]{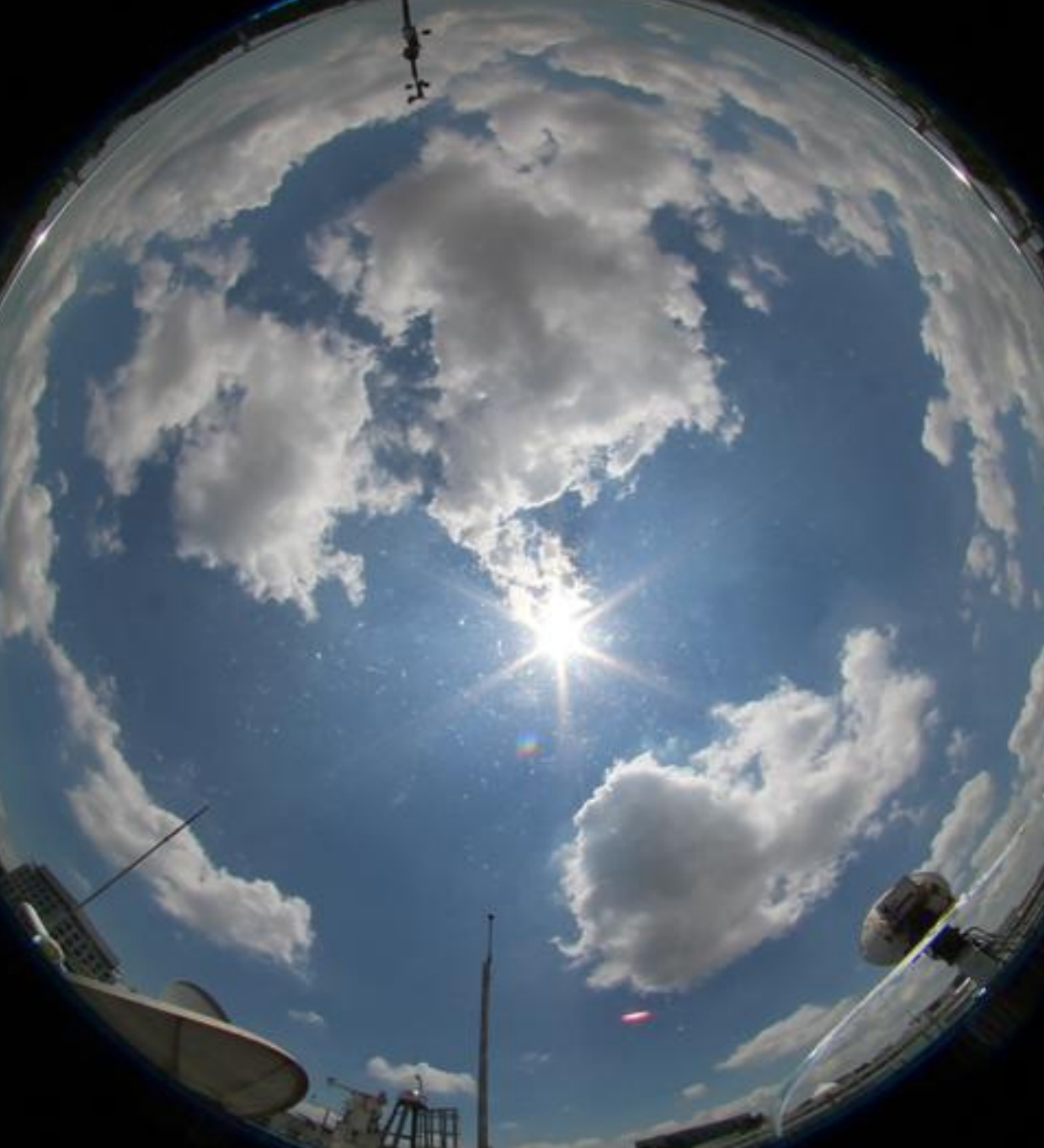}\\[2mm]
        \includegraphics[trim={6.9cm 2cm 7cm 1.8cm},clip,width=2.21cm]{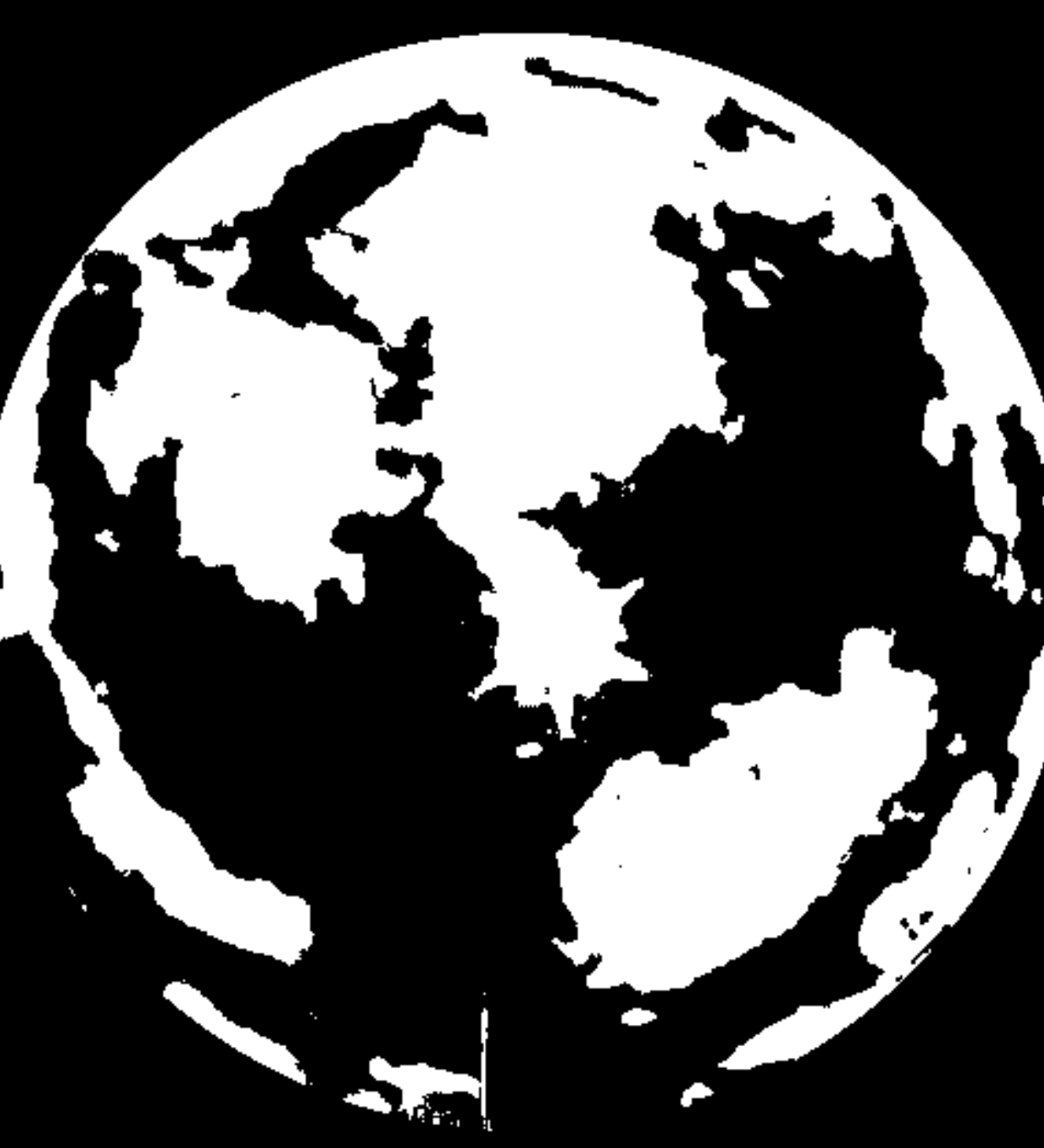}\\[2mm]
        \includegraphics[trim={6.9cm 2cm 7cm 1.8cm},clip,width=2.21cm]{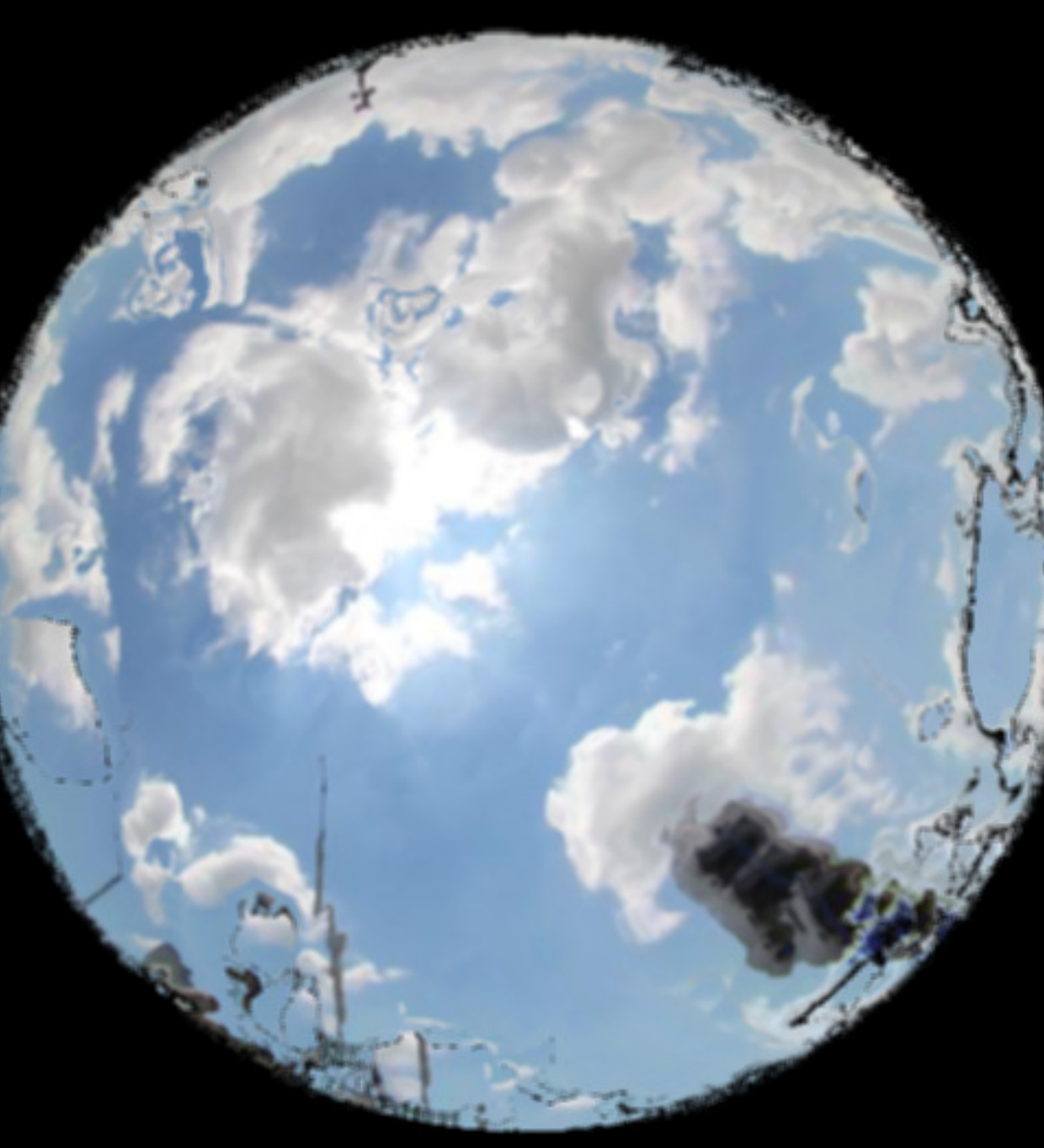}\\[2mm]
        \includegraphics[trim={6.9cm 2cm 7cm 1.8cm},clip,width=2.21cm]{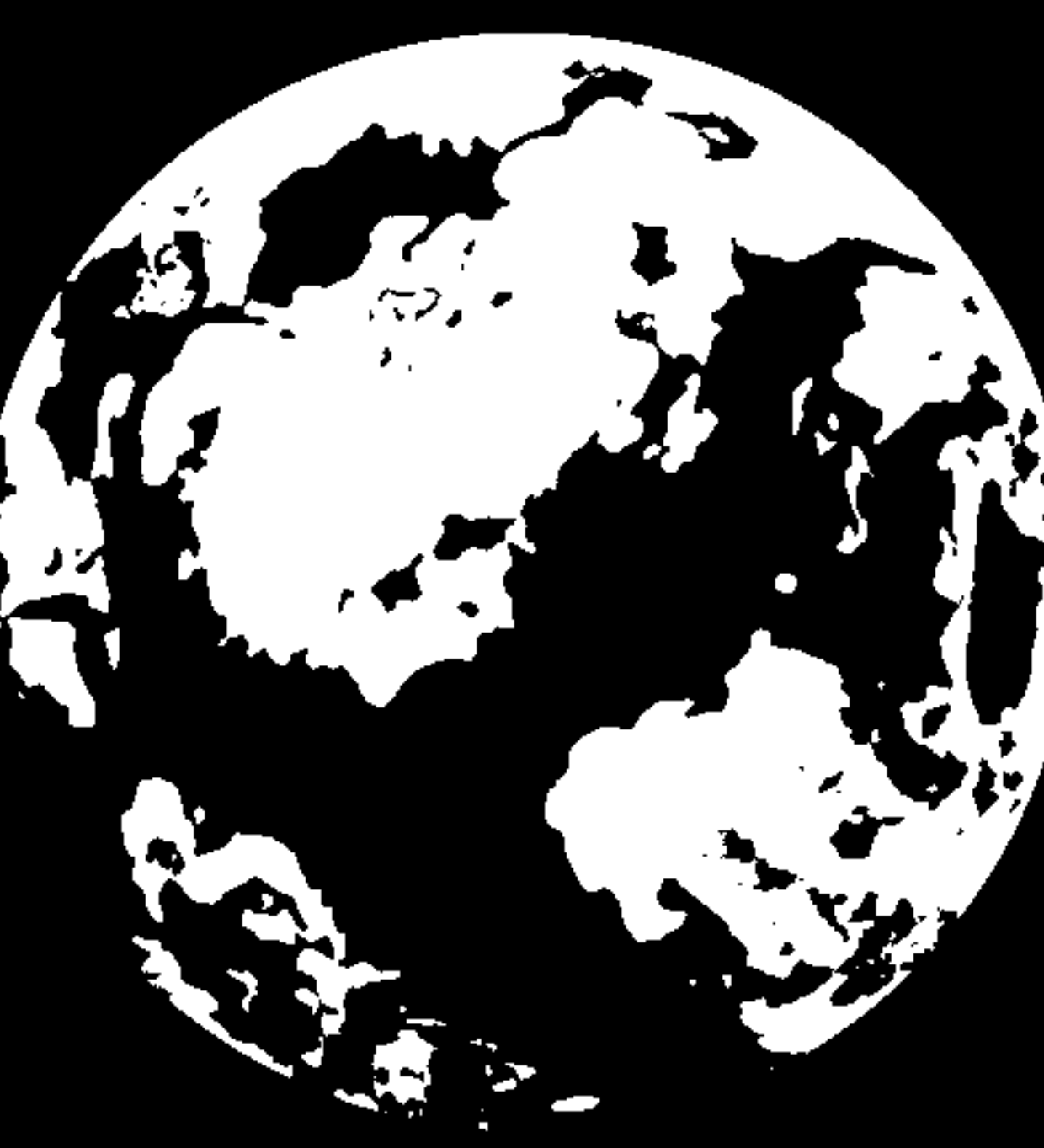}
        \caption{$t+4'$}
    \end{subfigure}
    \begin{subfigure}[t]{0.13\linewidth}
        \centering
        \includegraphics[trim={6.9cm 2cm 7cm 1.8cm},clip,width=2.21cm]{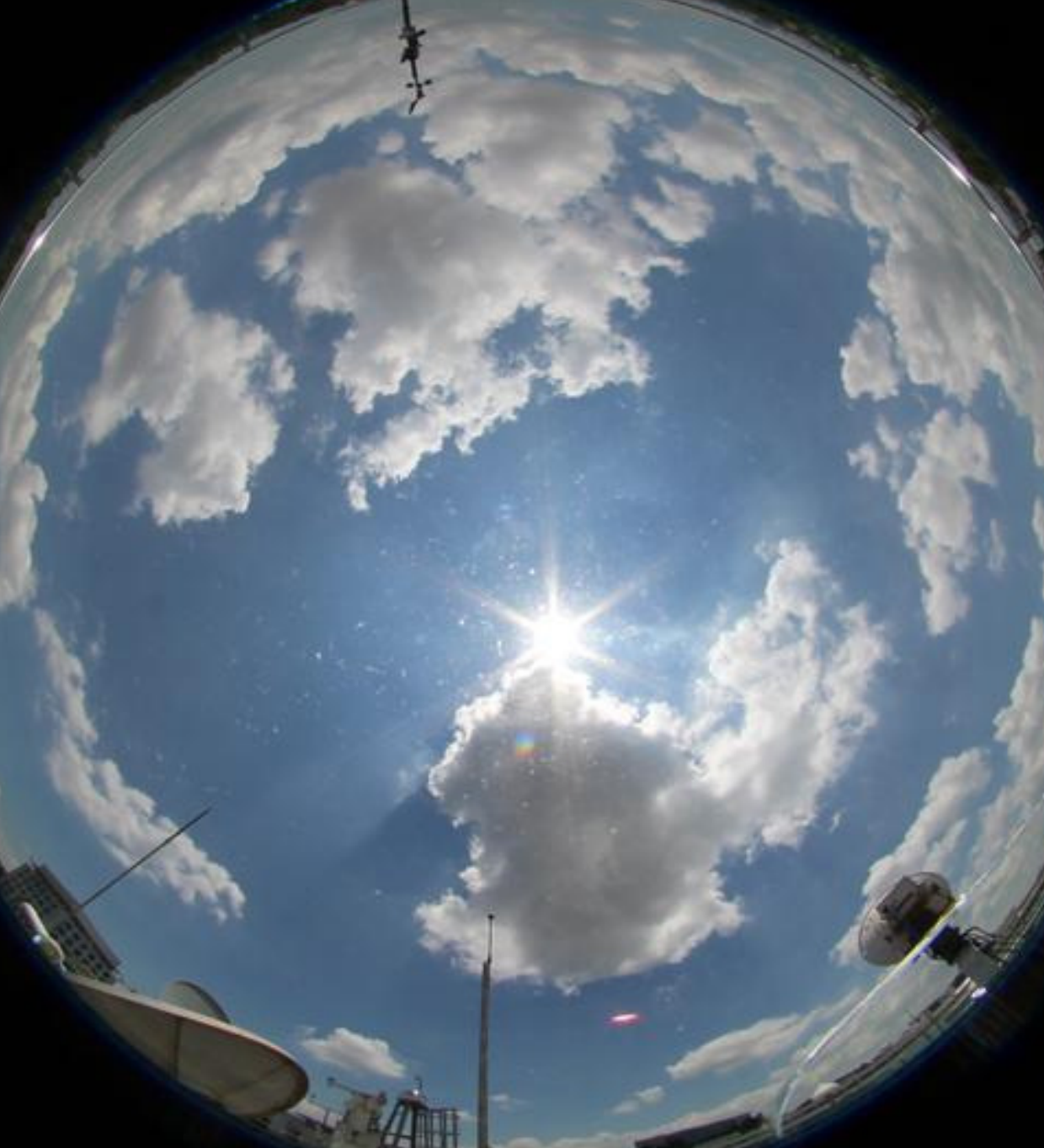}\\[2mm]
        \includegraphics[trim={6.9cm 2cm 7cm 1.8cm},clip,width=2.21cm]{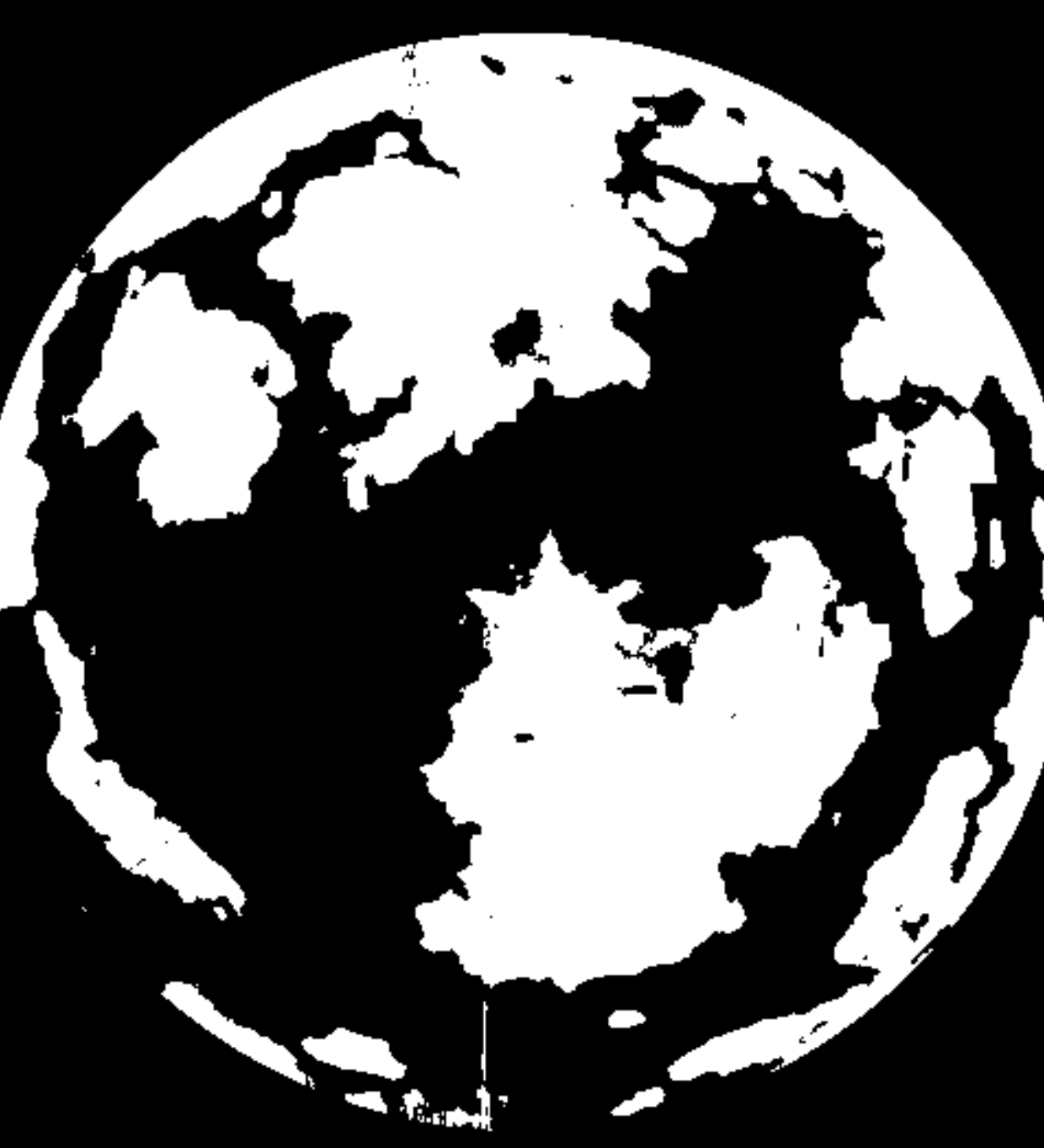}\\[2mm]
        \includegraphics[trim={6.9cm 2cm 7cm 1.8cm},clip,width=2.21cm]{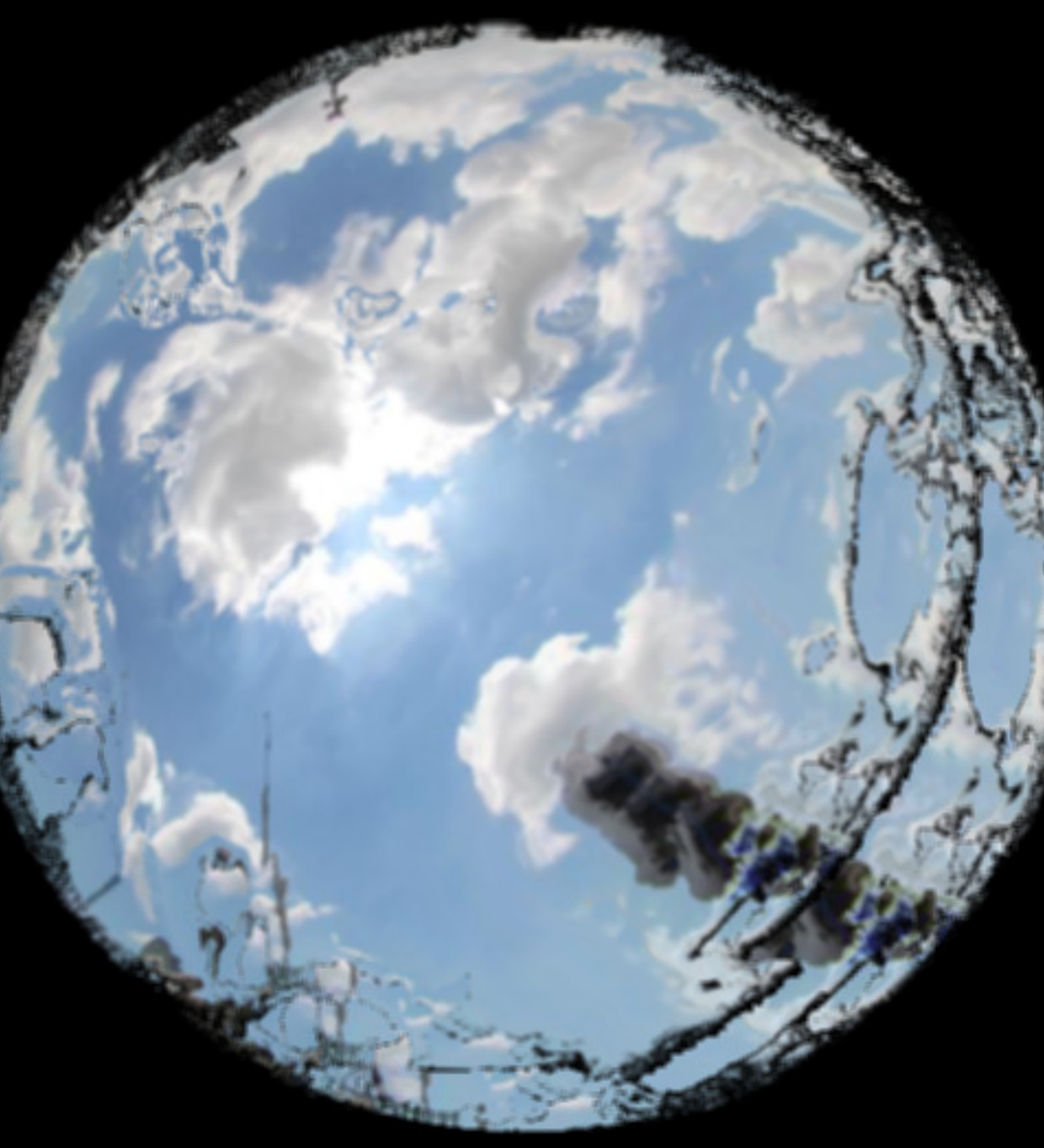}\\[2mm]
        \includegraphics[trim={6.9cm 2cm 7cm 1.8cm},clip,width=2.21cm]{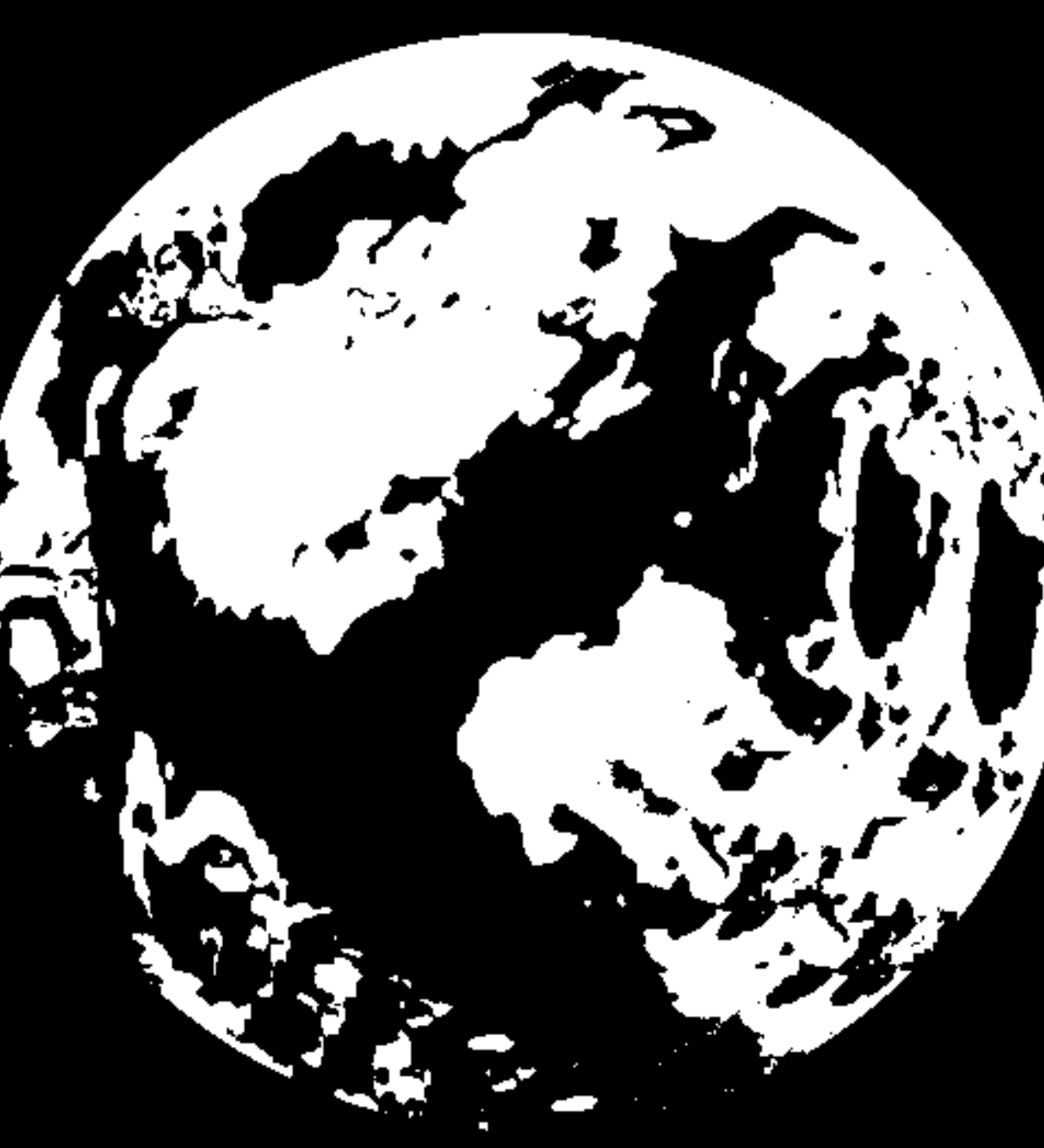}
        \caption{$t+6'$}
    \end{subfigure}
    \begin{subfigure}[t]{0.13\linewidth}
        \centering
        \includegraphics[trim={6.9cm 2cm 7cm 1.8cm},clip,width=2.21cm]{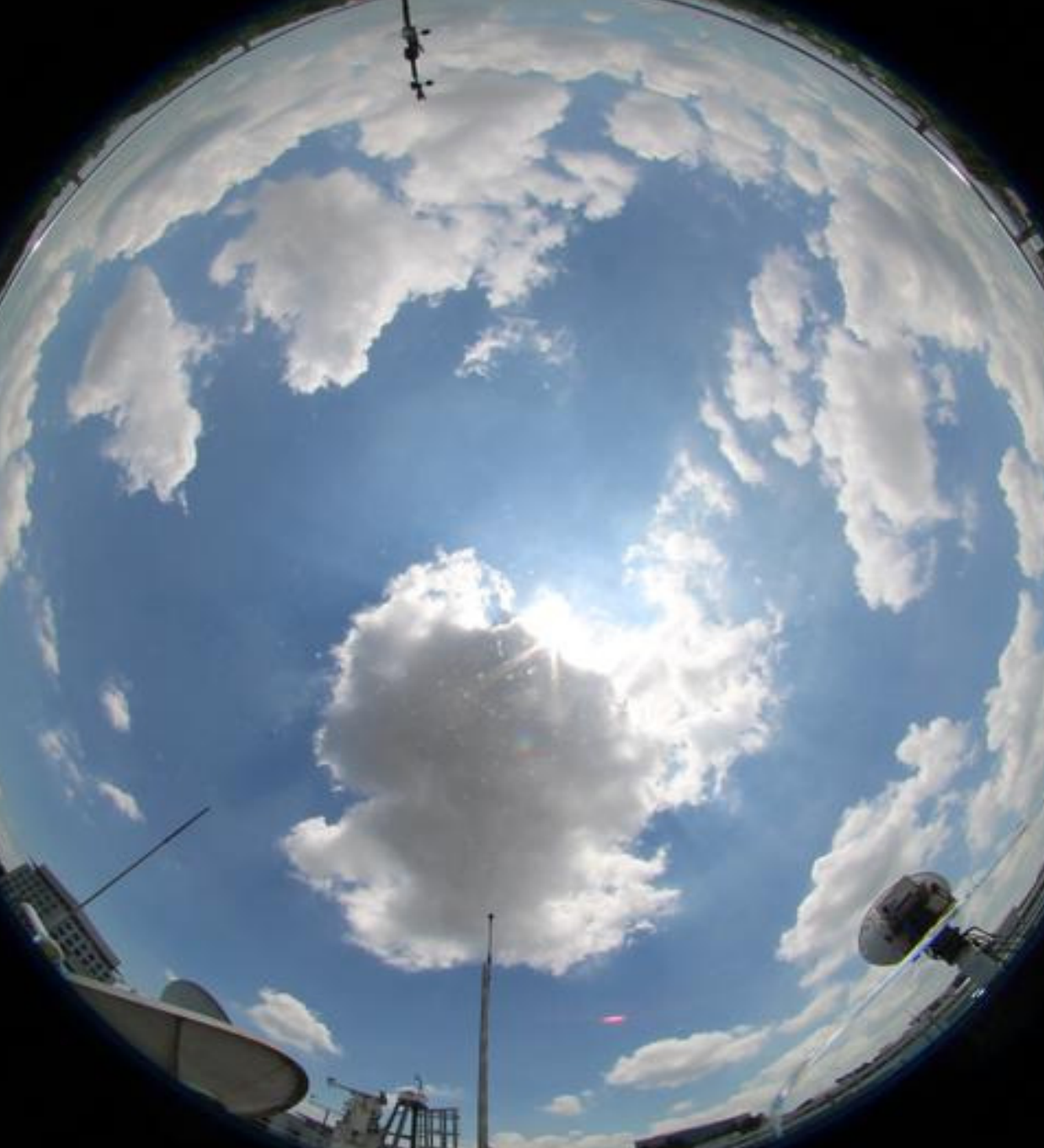}\\[2mm]
        \includegraphics[trim={6.9cm 2cm 7cm 1.8cm},clip,width=2.21cm]{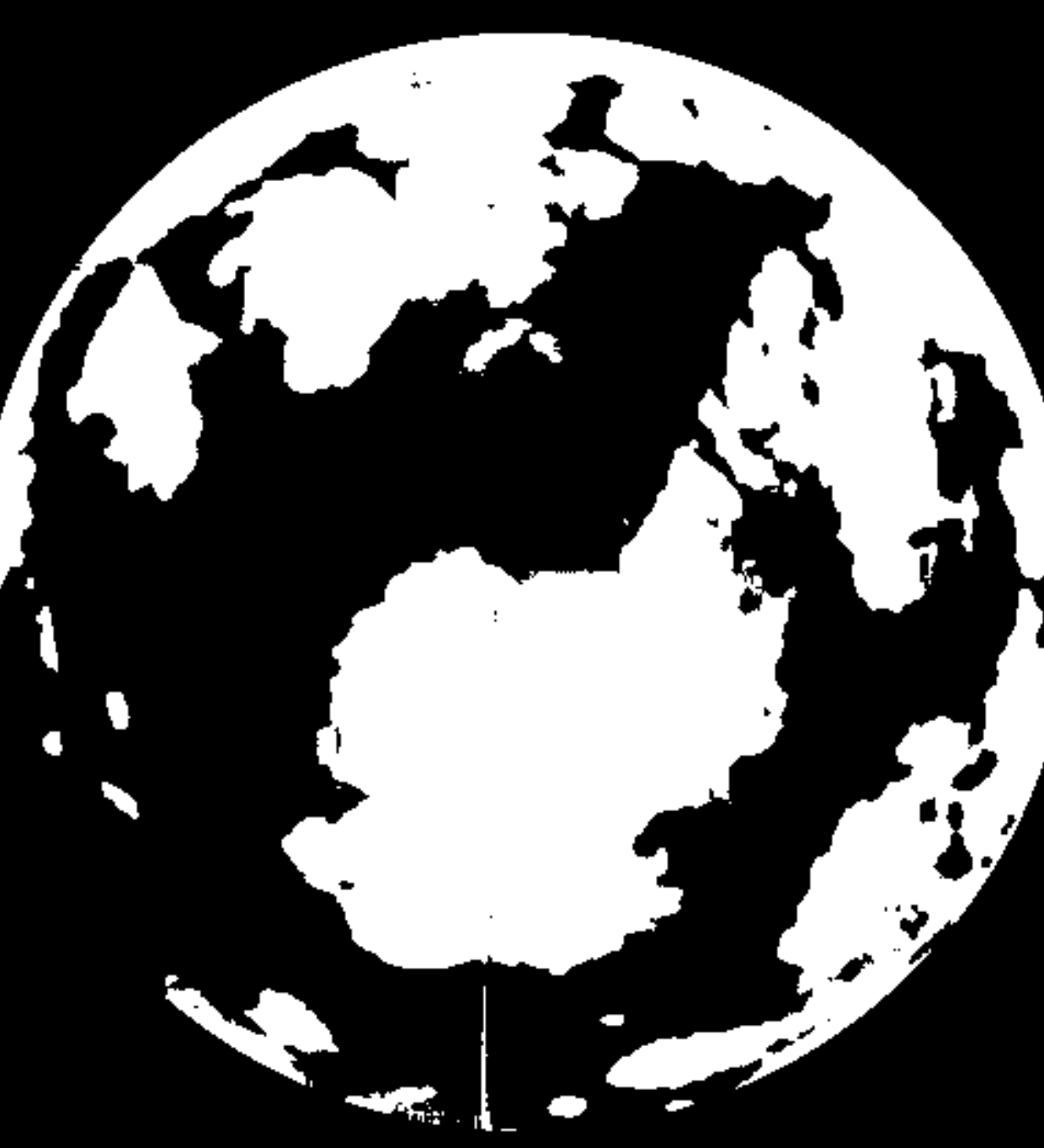}\\[2mm]
        \includegraphics[trim={6.9cm 2cm 7cm 1.8cm},clip,width=2.21cm]{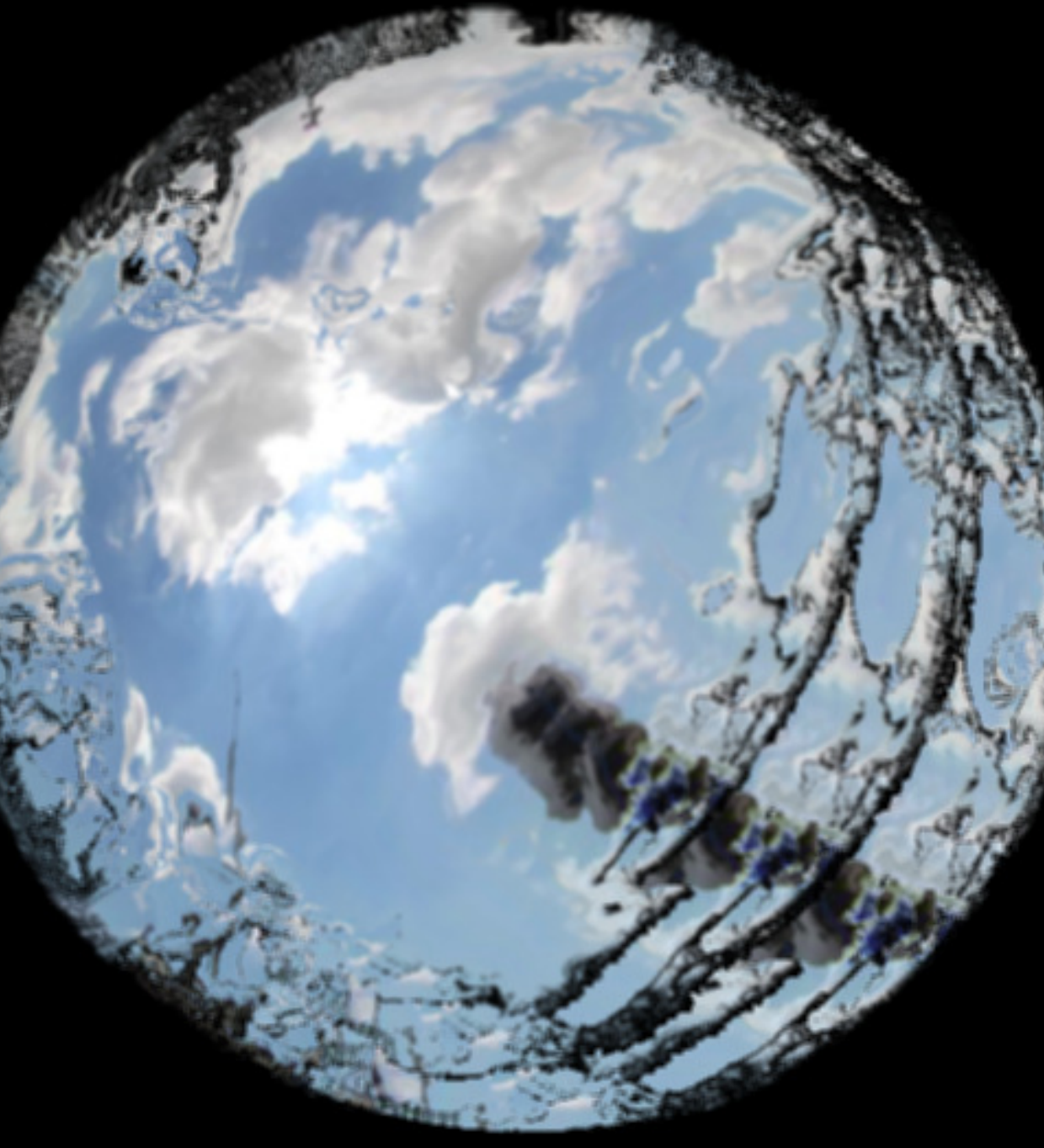}\\[2mm]
        \includegraphics[trim={6.9cm 2cm 7cm 1.8cm},clip,width=2.21cm]{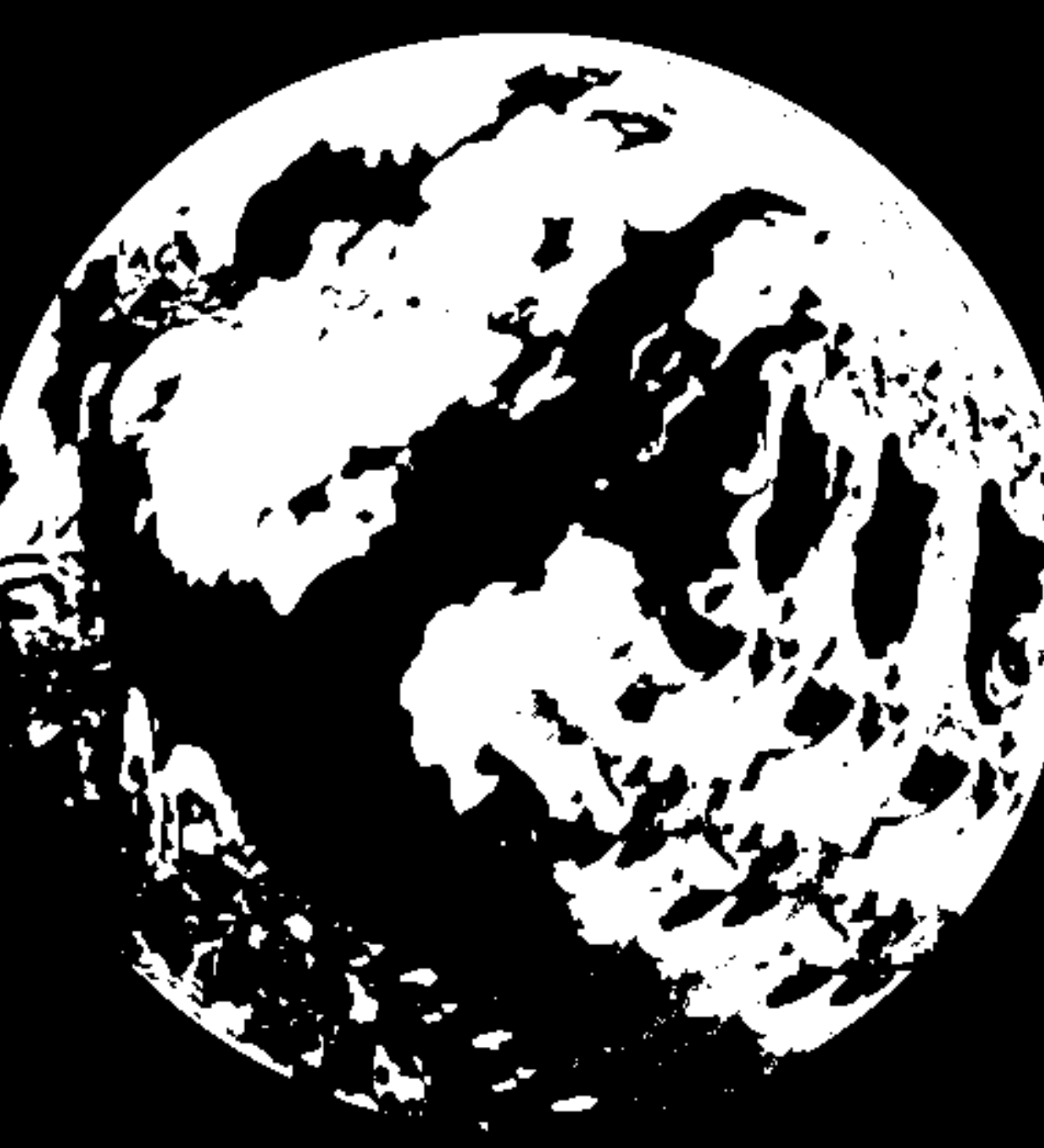}
        \caption{$t+8'$}
    \end{subfigure}
    \begin{subfigure}[t]{0.13\linewidth}
        \centering
        \includegraphics[trim={6.9cm 2cm 7cm 1.8cm},clip,width=2.21cm]{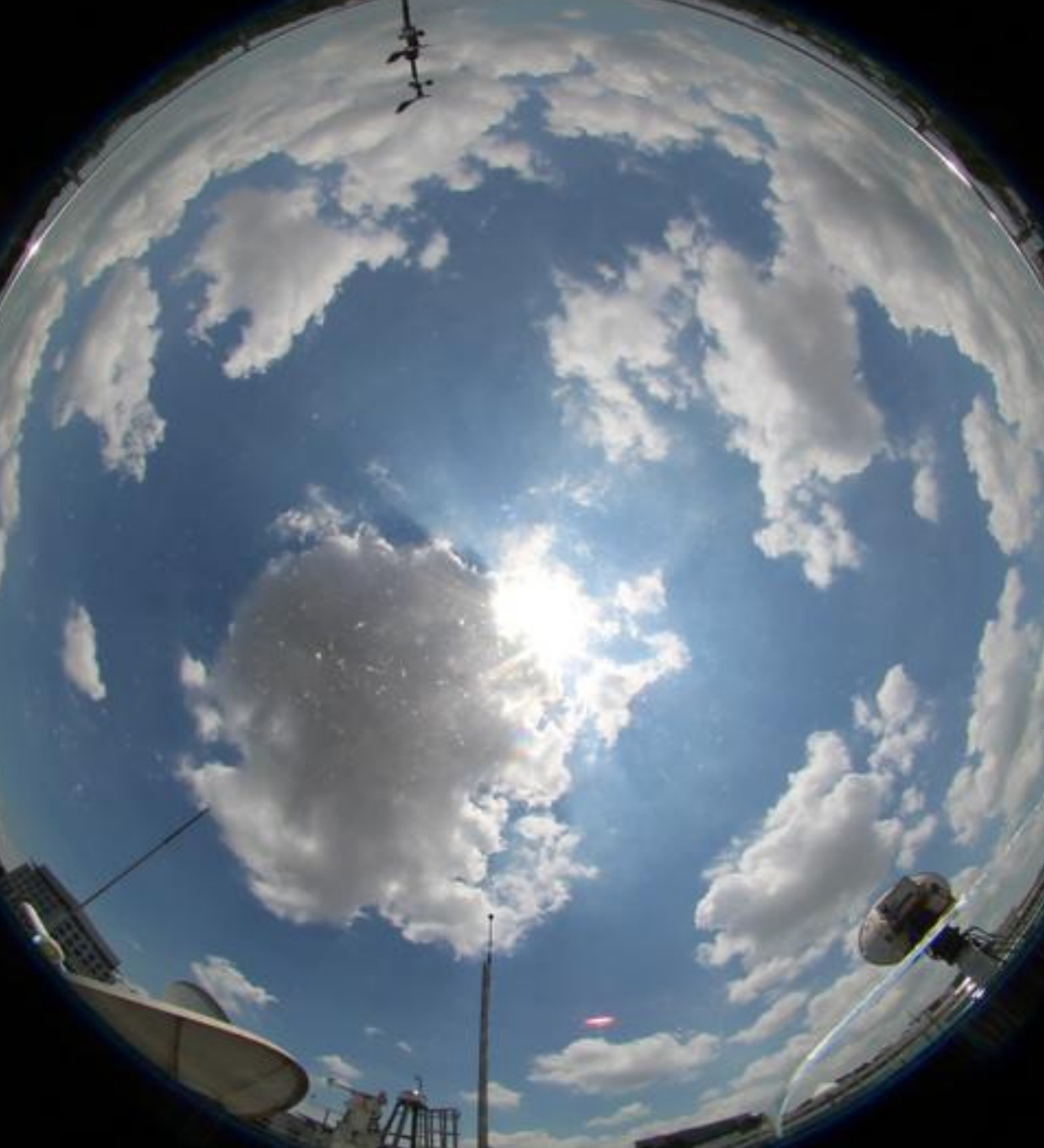}\\[2mm]
        \includegraphics[trim={6.9cm 2cm 7cm 1.8cm},clip,width=2.21cm]{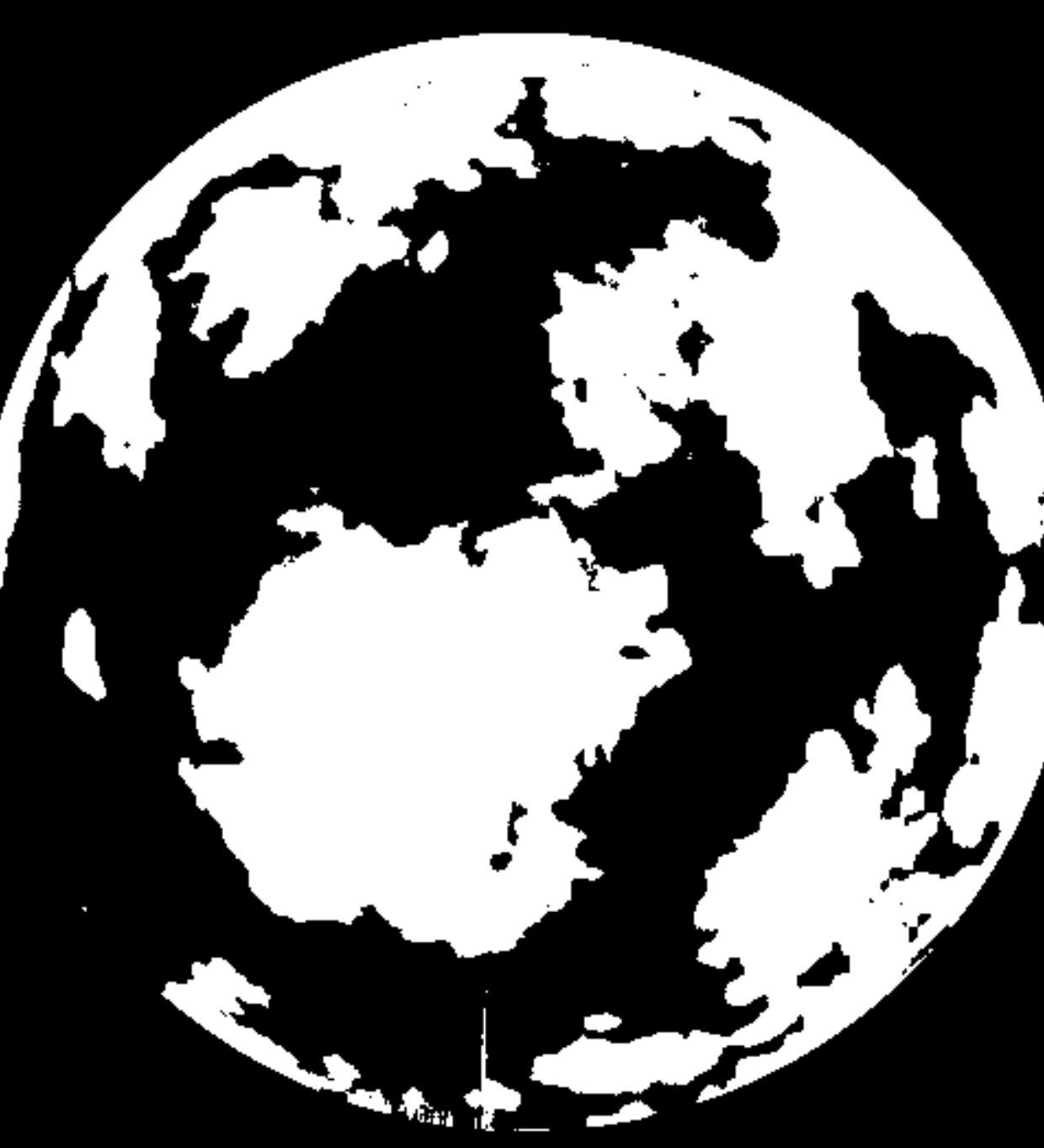}\\[2mm]
        \includegraphics[trim={6.9cm 2cm 7cm 1.8cm},clip,width=2.21cm]{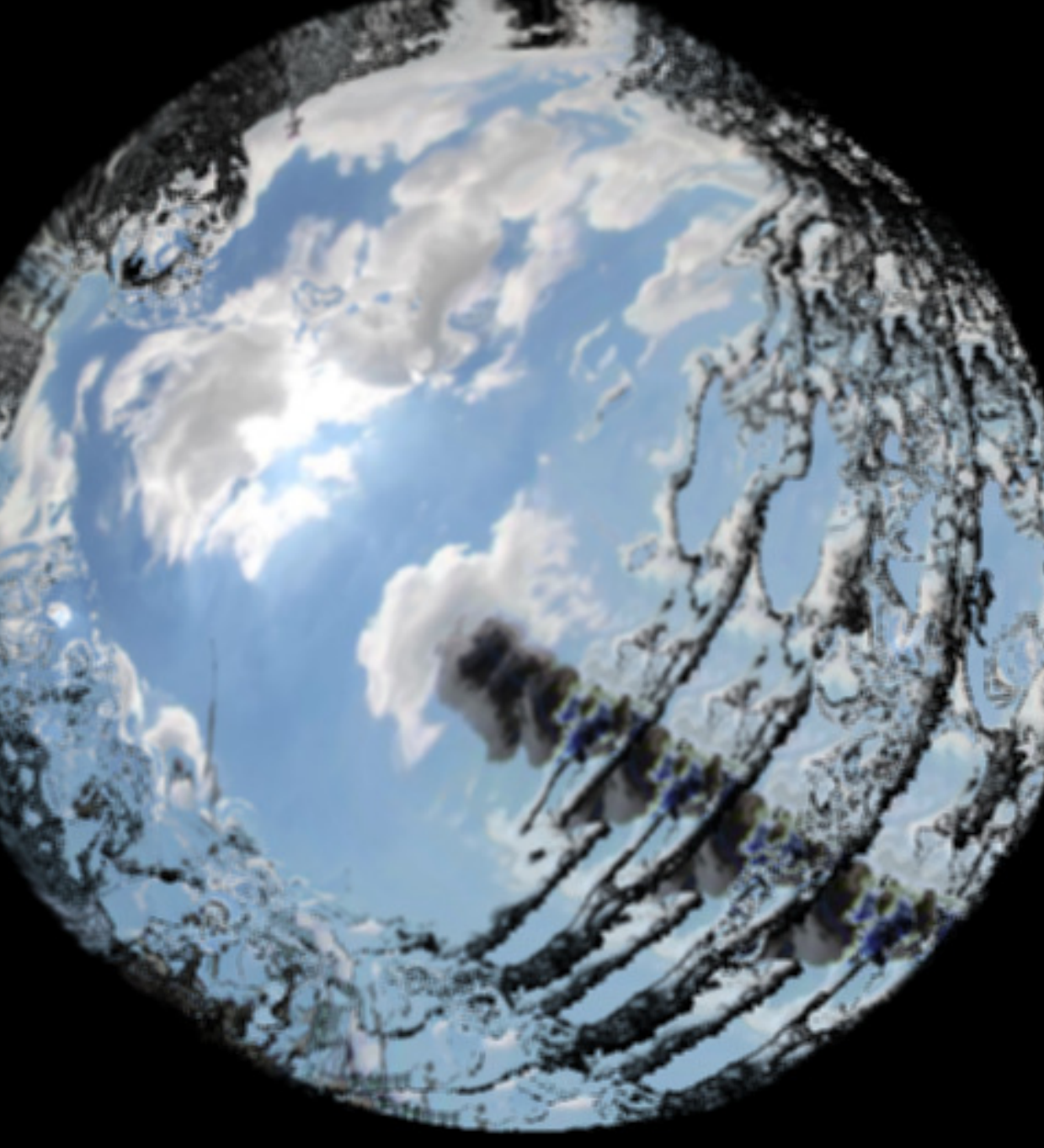}\\[2mm]
        \includegraphics[trim={6.9cm 2cm 7cm 1.8cm},clip,width=2.21cm]{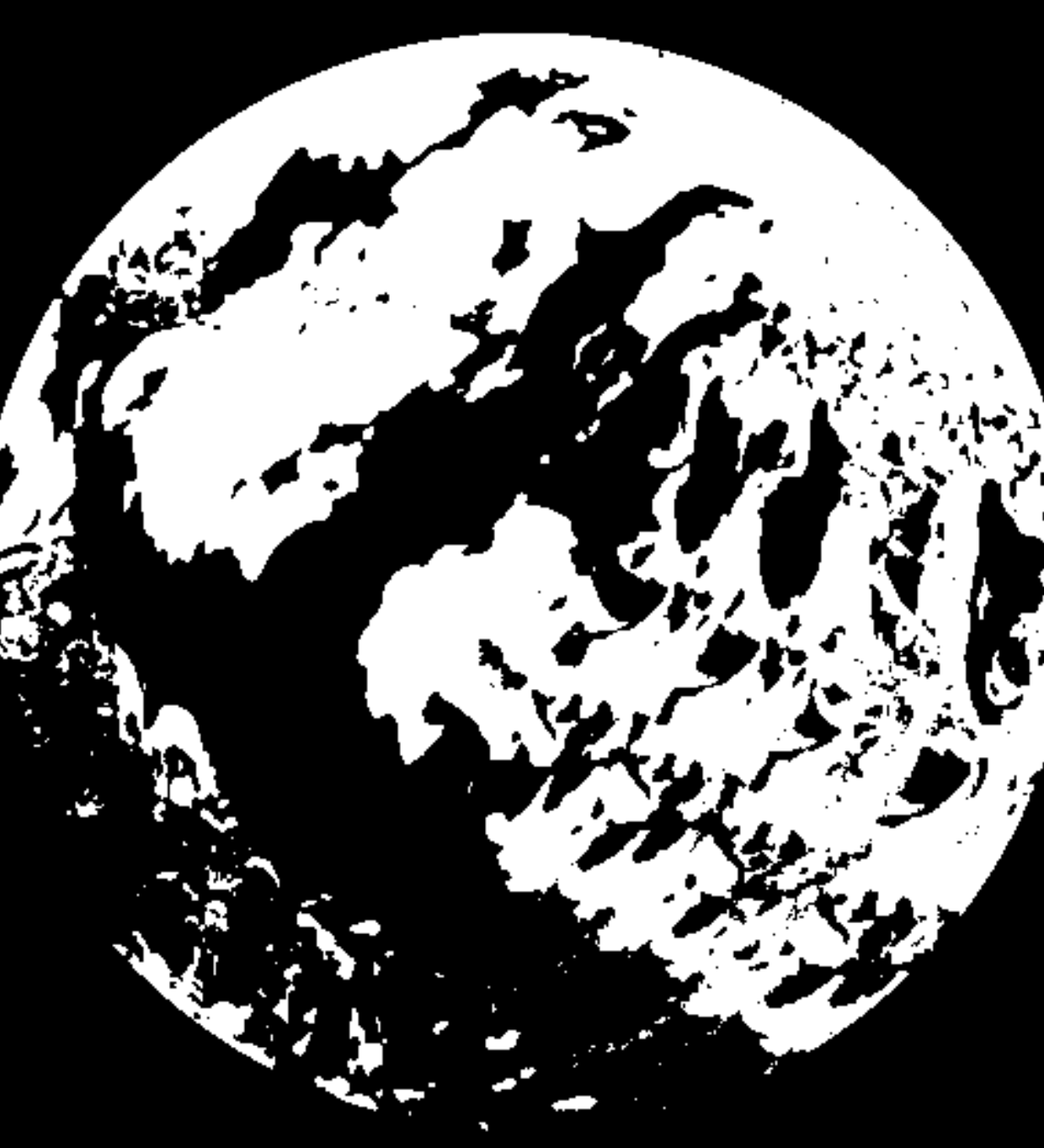}
        \caption{$t+10'$}
    \end{subfigure}
\caption{\label{fig:comb-example}Prediction of sky/cloud image on 16 April 2015 with a lead time of up to $10$ minutes, in intervals of $2$ minutes. The prediction accuracy is $83.44\%$ for a lead time of $2$ minutes, and $72.37\%$ for a lead time of $4$ minutes. The binary images are generated using our sky/cloud segmentation algorithm from \cite{ICIP1_2014}. We also observe that the actual image frame at time $t+10'$ significantly changes  compared to the actual image frame at time $t$.}
\end{figure*}

\section{Results \& Discussions}
\label{sec:results}
We now evaluate the forecasting accuracy of our methodology. For this purpose, we compute the binary sky/cloud image of the forecasted image using our cloud detection algorithm~\cite{ICIP1_2014}. We then compare it with the binary image computed from the original image. The accuracy is then calculated as the percentage of correctly classified pixels (sky or cloud) in the predicted binary image, as compared to the actual binary image. 

\begin{figure}[htb]
\begin{center}
\includegraphics[width=0.5\textwidth]{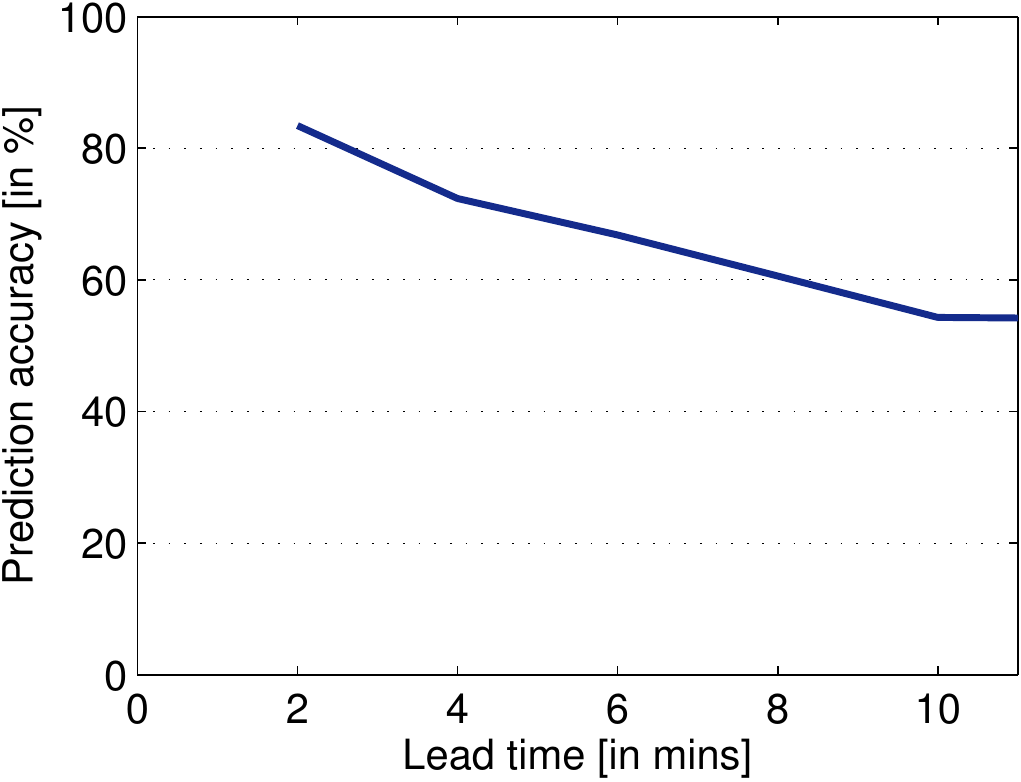}
\caption{Comparison of prediction accuracy percentage with different lead times.}
\label{fig:LT}
\end{center}
\end{figure}

We compute the prediction accuracy of our cloud tracking algorithm for a typical day in April 2015 and present our results for different lead times. Figure~\ref{fig:LT} shows the performance of our cloud tracking algorithm for different lead times.  We observe that prediction accuracy is good for short lead times, but gradually decreases as lead times become larger. This makes sense as clouds generally move quite fast, and can change shape between image frames.  Furthermore, the error obtained in the intermediate forecast image is cascaded to future images.

We provide a few illustrative examples of our prediction accuracy upto $10$ minutes. Figure~\ref{fig:comb-example} shows the output of our algorithm with a lead time of upto $10$ minutes, in intervals of $2$ minutes. %In (c) to (f), 
We use the frame at time \emph{t-2} minutes and the frame at time \emph{t} as the input images of our proposed approach to predict the frame at time \emph{t+2} minutes. We show the actual and predicted images along with their corresponding binary images. The accuracy achieved with our algorithm is $83.44\%$. We then use the frame at \emph{t} minutes along with the predicted image at \emph{t+2} minutes to predict the frame at \emph{t+4} minutes. %, which we show in (g) to (j). 
The accuracy is still good (around $72.37\%$) given the higher lead time. However, we observe for higher lead times, the predicted image gets progressively more distorted as compared to its actual image, and artifacts appear. This happens because the error incurred in the previous lead times gets cascaded to future frames, as clouds significantly changes in shape over a relatively short period of time.

We achieve a prediction accuracy above 70\% for lead times of 4 to 6 minutes. It however gradually decreases with longer lead time. After 10 minutes, the clouds have moved significantly, and there is little correlation between frames, as shown in Fig.~\ref{fig:comb-example}. These predictions are only based on images capturing a small area of the sky, and better long-term predictions would only possible at a larger scale and a lower level of detail.

\section{Conclusions}
\label{sec:concl}
In this paper, we have discussed about our methodology to track cloud movement across successive image frames from sky cameras. It is based on optical flow and performs well for lead times of a few minutes. The accuracy gradually decreases for larger lead times. Our proposed approach is intended mainly for the short-term prediction of cloud movements, as we perform a localized analysis of cloud motion. In our future work, we plan to use other meteorological data, such as wind sensors, to further increase the prediction accuracy.

\balance

\bibliographystyle{IEEEbib}

\end{document}